\documentclass[runningheads]{llncs}
\usepackage[T1]{fontenc}
\usepackage{graphicx}
\usepackage{times}
\usepackage{amsmath}
\usepackage{amssymb}
\usepackage{algorithm}
\usepackage{algorithmic}

\begin{document}
\title{Deep reinforcement learning for weakly coupled MDP's with continuous actions}
%
%
\author{Francisco Robledo\inst{1}\orcidID{0000-0003-1040-1513} \and
Urtzi Ayesta\inst{2}\orcidID{0000-0003-1761-2313} \and
Konstantin Avrachenkov\inst{3}\orcidID{0000-0002-8124-8272}}
\authorrunning{F. Robledo et al.}
%
\institute{UPV/EHU, Univ. of the Basque Country, 20018 Donostia, Spain \\ UPPA, Université de Pau et des Pays de l'Adour, 64000 Pau, France
\email{frrobledo96@gmail.com} \and
IRIT, Université de Toulouse, CNRS, Toulouse INP, UT3, Toulouse, France\\ UPV/EHU, Univ. of the Basque Country, 20018 Donostia, Spain\\ IKERBASQUE - Basque Foundation for Science, 48011 Bilbao, Spain
\email{urtzi.ayesta@irit.fr} \and
INRIA Sophia Antipolis, France 
\email{k.avrachenkov@inria.fr}}
\maketitle              
\begin{abstract}
This paper introduces the Lagrange Policy for Continuous Actions (LPCA), a reinforcement learning algorithm specifically designed for weakly coupled MDP problems with continuous action spaces. LPCA addresses the challenge of resource constraints dependent on continuous actions by introducing a Lagrange relaxation of the weakly coupled MDP problem within a neural network framework for Q-value computation. This approach effectively decouples the MDP, enabling efficient policy learning in resource-constrained environments. We present two variations of LPCA: LPCA-DE, which utilizes differential evolution for global optimization, and LPCA-Greedy, a method that incrementally and greadily selects actions based on Q-value gradients. Comparative analysis against 
other state-of-the-art techniques 
across various settings highlight LPCA's robustness and efficiency in managing resource allocation while maximizing rewards.


\end{abstract}
\section{Introduction}\label{sec: introduction}
The exploration of optimal decision-making under uncertainty is a fundamental problem \cite{sutton2018reinforcement}, with significant implications in diverse fields such as telecommunications, finance, robotics, and healthcare. At the heart of this exploration lies the restless multi-armed bandit (RMAB) problem, an extension of the classical multi-armed bandit framework \cite{Gittins1979} to scenarios where arms evolve independently of the player's actions. Introduced by \cite{whittle1988restless}, the RMAB problem highlights the challenge of allocating limited resources among competing projects or processes in a state of continuous change. Recently, many studies have focused on neural network approximation in restless bandit problems, such as the works of \cite{Avrachenkov2022}, \cite{Robledo2022}, and \cite{Nakhleh2021}, which use deep reinforcement learning to approximate the Whittle indices used in their heuristics.

One can generalize the restless bandits to weakly coupled MDPs, where the independent MDPs are coupled only through a constraint on the action and actions can belong to complex spaces. These problems present substantial complexity due to constraints of the actions and common resources. 
A key advancement in addressing such complex problems came with the introduction of Lagrangian Decomposition methods, as explored by \cite{hawkins2003langrangian}. The approach of \cite{hawkins2003langrangian} proposes a Lagrangian decomposition approach for solving the weakly coupled dynamic optimization problem, which yields upper bounds as well as heuristic solutions.
Works by \cite{Srinivas2012} and \cite{Meshram2020} have introduced methods for navigating these complex decision spaces, employing Gaussian processes and simulation-based algorithms, respectively, to tackle the multi-action challenges.

Other studies in weakly coupled MDPs include the work of \cite{Wei2018}, which addresses the challenges of online learning in this specific MDP setting and presents an algorithm with a tight $O(\sqrt{t})$ regret and constraint violations simultaneously. Additionally, \cite{Gast2022a} introduces the LP-update policy, which generalizes the classical restless bandit problems and demonstrates asymptotic optimality at various rates depending on problem characteristics. 
 
Significant advances in deep reinforcement learning include the development of Deep Deterministic Policy Gradient (DDPG) \cite{Lillicrap2019} and Twin Delayed DDPG (TD3) \cite{Fujimoto2018}, algorithms that have significantly advanced complex control tasks by solving MDPs with continuous actions. Building on the capabilities of these frameworks, the OptLayer architecture was introduced \cite{Pham2018}, specifically designed to generate safe, constraint-compliant actions. OptLayer integrates an additional layer that solves a constraint optimization problem applicable to both DDPG and TD3 architectures. This extension ensures that the actions taken by the learning models adhere to predefined constraints. \cite{Killian2021} explores the online learning landscape for discrete multi-action RMABs and presents a Q-learning Lagrange policy algorithm tailored for restless multi-armed bandits with multiple discrete actions. 
Similarly, \cite{ElShar2023} uses this Lagrangian decomposition to train separate subagents for each individual MDP problem, and a general network to combine these results, also in the context of discrete multi-action RMABs. 

In this work, we introduce the Lagrange Policy for Continuous Actions (LPCA) algorithm, a reinforcement learning algorithm specifically designed for weakly coupled MDP problems with continuous action spaces. To the best of our knowledge, this is the first paper proposing an algorithm to solve weakly coupled MDPs with continuous actions. 
LPCA integrates a neural network-based framework to study weakly coupled MDP using the Lagrange relaxation introduced in \cite{hawkins2003langrangian} to decouple the projects of the MDP, being able to study their dynamics independently of one another and effectively balancing resource constraints and individual project decisions. 
Continuous actions allow for a more accurate representation of real-world scenarios, such as adjusting resource levels or control parameters, without the limitations of discretization. This flexibility enhances the algorithm's ability to optimize performance by better managing trade-offs between competing processes, ultimately leading to more robust and efficient policy learning.

\section{Problem Formulation}\label{sec: problem_formulation}
In our approach to the weakly coupled MDPs with continuous actions, we consider an environment consisting of $N$ projects, each characterized by its unique state, action, and the resulting reward. Specifically, the state of 
the system is given by $\mathbf{s}=(s_1,...,s_N) \in \mathbf{S}$, where each project is represented as $s_i$, an element from the finite state space $S_i$, $i=1,...,N.$ Correspondingly, the actions taken in each project are denoted as elements $ a_i $ belonging to the compact action space $A_i $, and the complete system action is denoted with bold font $\mathbf{a}=(a_1,...,a_N) \in \mathbf{A}$. The rewards obtained from these actions are encapsulated as elements $ r_i $ in the reward vector $ \mathbf{r}$. The cost associated with each action $ a_i $ is expressed as $ c(a_i) $, and the cumulative cost for all actions is given by $ C(\mathbf{a}) = \sum_i c(a_i) $.

The system dynamics are governed by a transition probability kernel $T: \mathbf{S} \times \mathbf{A} \times \mathbf{S} \rightarrow [0,1]$, which specifies the probabilities of transitioning to new states given particular state and action vector. Given the values of actions, $T$ has a product form. A discount factor $\gamma \in (0,1)$ is used to balance immediate and future rewards.

The long-term discounted reward can be expressed through the Bellman value function $V(\mathbf{s})$, which is the expected sum of discounted rewards accumulated over time, starting from the state $\mathbf{s}$ and satisfying the Bellman dynamic programming equation:
\begin{equation}
\label{eq: bellman coupled}
V(\mathbf{s}) = \max_{\mathbf{a} \in \mathbf{A}, \, C(\mathbf{a}) = B}\left[ \sum_{i=1}^{N}r_i(s_i,a_i) + \gamma \mathbb{E}[V(\mathbf{s}') \mid \mathbf{s}, \mathbf{a}]\right].   
\end{equation}
The complexity of the problem comes primarily from the constraint imposed on the actions, which are dictated by a common pool of resources. Specifically, each project must select a continuous action $a_i \in [0, a_i^{\max}]$ whose activation cost, represented by the total cost $C(\mathbf{a})$, directly consumes a predefined total pool of available resources $B$. This shared resource pool constraint means that actions across projects are inherently coupled, which significantly increases the complexity of the decision space as the number of projects increases. The exponential growth in decision space complexity due to this coupling underscores the challenge of resource allocation and emphasizes the need for efficient use of the shared resource pool \cite{Bertsekas2012}.

To manage this complexity, we can relax the value function using a Lagrange multiplier $ \lambda $. This transforms the original problem into a Lagrangian form:

\begin{equation}\label{eq: decoupled value function}
    \begin{aligned}
        J(\mathbf{s}, \lambda) =
        & \max_{\mathbf{a} \in \mathbf{A}}\left[\sum_{i=1}^N r_i(s_i,a_i) + \lambda \left(B - \sum_{i=1}^N c(a_i) \right) + \gamma \mathbb{E}[J(\mathbf{s'}, \lambda) \mid \mathbf{s}, \mathbf{a}] \right].
    \end{aligned}
\end{equation}

Here, $\lambda$ is the Lagrange multiplier associated with the resource constraint $ B $. By adjusting $ \lambda $, we effectively balance the immediate cost of actions against their long-term rewards, allowing for a decoupling of the projects' decisions. 
If we assume the additive structure of the value function with respect to the projects of the weakly coupled MDP, the equation (\ref{eq: decoupled value function}) can be rewritten as:
\begin{equation}\label{eq: J-vector}
J(\mathbf{s}, \lambda) = \frac{\lambda B}{1-\gamma} + \sum_{i=1}^N \max_{a_i \in A_i} Q_i(s_i, a_i, \lambda),
\end{equation}
where
\begin{equation}\label{eq: Q-value definition}
    Q_i(s_i, a_i, \lambda) = r_i(s_i,a_i) - \lambda c(a_i) + \gamma \sum_{s'_i}T(s_i, a_i, s'_i) \max_{a_i' \in A_i}Q_i(s'_i, a'_i, \lambda).  
\end{equation}
In this decoupled framework, the Lagrange multiplier $\lambda$ is instrumental in determining the optimal policy for each project. Under the budget constraint $B$, $\lambda$ acts as a trade off parameter by introducing a penalty term $\lambda c(a_i)$ for the actions taken. 
A higher $\lambda$ parameter places more emphasis on minimizing the cost (i.e., staying within the resource limit $B$), while a lower $\lambda$ value shifts the focus towards maximizing rewards with less emphasis on the cost implementations. 
As $\lambda$ rises, the preferred policy for each project will increasingly favor actions that offer the highest ``value-to-cost'' ratio.
Thus, the function \eqref{eq: J-vector} is a measure of the total expected reward, adjusted for the cost of the actions taken under that policy.
To balance the expected rewards with the cost of actions, we need to find $\lambda^*$ such that
\begin{equation}\label{eq: lambda optimal}
    \lambda^*(\mathbf s) = \arg\min_\lambda J(\mathbf{s}, \lambda).
\end{equation}

This term is defined as the best trade-off between maximizing rewards and minimizing the cost of actions. It is at this point that the policy aligns with the time-averaged resource constraints, ensuring that the actions selected are not only rewarding but also resource-efficient.

Then, in a continuous action framework, at each time step $t$  
we aim to solve the following Knapsack-like optimization problem:
\begin{equation}\label{eq: original constraint - forced}
    \begin{aligned}
    & \max_{\mathbf{a} \in \mathbf{A}} && \sum_{i=1}^{N} Q_i(s_i(t),a_i,\lambda^*(s_i)) \quad \quad  s.t. \quad \sum_{i=1}^{N} c(a_i) = B.
\end{aligned}
\end{equation}
In the LPCA algorithm, described in detail next, we 
interpolate the curve of the Q-values $Q(s,a,\lambda)$ as functions of the Lagrange multiplier $\lambda$ through a neural network. This curve is a convex function with respect to $\lambda$ \cite{hawkins2003langrangian}, making the  minimization of  \eqref{eq: J-vector} a simple 
one-dimensional convex optimization problem once the neural network is trained. 
For the optimization (\ref{eq: original constraint - forced})
we explore two approaches as outlined in Sections~\ref{sec:differential evolution}~and~\ref{sec: greedy}.

\section{LPCA Algorithm}\label{sec: algorithm_description}
\begin{algorithm}[t]
    \caption{LPCA Training Process}\label{alg: General LPCA}
    \begin{algorithmic}[1]
        \REQUIRE Environment, $N_{\text{iter}}$, Update frequency $N$, Batch size $M$, Policy method
        \ENSURE Train LPCA Model, Update Policy Dictionary
        \STATE Initialize Q-value neural network, policy dictionary, experience memory
        \FOR{$\text{iteration} = 1$ \TO $N_{\text{iter}}$}
            \STATE Select and execute action $\mathbf{a}$, store ($\mathbf{s}, \mathbf{a}, \mathbf{r}, \mathbf{s'}$, done)
            \IF{$\text{memory} \geq M$}
                \STATE Update Q-values with mini-batch of $M$ (Algorithm \ref{alg: Update Q-values})
            \ENDIF
            \IF{$\text{iteration} \mod N = 0$}
                \STATE Update policy with Differential Evolution or Greedy (Algorithm \ref{alg: policy update})
            \ENDIF
        \ENDFOR
    \end{algorithmic}
    \end{algorithm}
In numerous practical applications, the model parameters, particularly expected rewards and transition probabilities, are often unknown or inaccessible. To address this, traditional reinforcement learning methods have been employed to learn those parameters \cite{sutton2018reinforcement}. 
However, a significant challenge arises in environments where the projects of the MDP are coupled. In these cases, the complexity of solving the problem increases exponentially with the number of projects.
To address this challenge, we introduce LPCA, a reinforcement learning algorithm that extends Q-learning by incorporating neural networks for approximating Q-values for constrained continuous actions. This section details the operation and implementation of LPCA.

The core methodology of the LPCA algorithm involves a two-timescale process centered around learning and optimization. Initially, LPCA focuses on training a neural network to accurately approximate the Q-values as defined in Equation \eqref{eq: Q-value definition}. This process involves learning the balance between immediate rewards, action costs, and future rewards based on the transition dynamics of the system. Once the neural network is effectively trained, in online fashion, for the current coupled state $ \mathbf{s} $, LPCA computes the value function $ J(\mathbf{s}, \lambda) $ as described in Equation \eqref{eq: J-vector}. The objective is to determine the optimal Lagrange multiplier $ \lambda^* $ that minimizes $ J(\mathbf{s}, \lambda) $ as formulated in Equation \eqref{eq: lambda optimal}. Finally, LPCA addresses the optimization problem set out in Equation \eqref{eq: original constraint - forced} through two possible methods: a differential evolution optimizer (Algorithm \ref{alg:differential_evolution}) or a greedy optimizer (Algorithm \ref{alg:greedy_selection}).

The general training process of LPCA, as outlined in Algorithm \ref{alg: General LPCA}, is a key aspect of our approach. The algorithm begins by utilizing a policy dictionary to interact with the environment.
This dictionary is a mapping of states to actions, where each state corresponds to a unique action vector. 
During each interaction, an action is selected based on the current policy, and the environment responds accordingly. The response, including the state transition and reward information, is stored as a transition sample. Notably, each process of the weakly coupled MDP is treated individually, with the transition sample from each project recorded separately in a memory buffer. This memory serves as a repository for experiences, which are later used to update the neural network that approximates Q-values.

\begin{algorithm}[t]
    \caption{Update Q-values in LPCA Neural Network Model}\label{alg: Update Q-values}
    \begin{algorithmic}[1]
        \FOR{each random sample in memory}
            \STATE Extract $s$, $a$, $r$, $s'$, $is\_terminal$ from sample \COMMENT{$is\_terminal$ indicates if $s'$ is a terminal state}
            \STATE $Q \leftarrow$ Calculate target Q-values for $s$ and $a$ using a subset of $\lambda$ values lambda\_grid
            \STATE $V_{expected} \leftarrow$ Calculate expected value functions for $s'$ using target network for each $\lambda \in$ lambda\_grid
            \IF{$is\_terminal$}
                \STATE $Q_{target}(s,a,\lambda) \leftarrow r(s) - \lambda c(a)$
            \ELSE
                \STATE $Q_{target}(s,a,\lambda) \leftarrow r(s) - \lambda c(a) + \gamma \cdot V_{expected}$
            \ENDIF
            \STATE Perform a gradient descent step on $(Q_{target}(s,a, \lambda) - Q(s, a, \lambda))^2$ to update network weights
        \ENDFOR
        \STATE Perform soft-update on target network weights $\theta' \leftarrow \theta \tau + (1-\tau) \theta'$
    \end{algorithmic}
\end{algorithm}

The training of the neural network, as detailed in Algorithm \ref{alg: Update Q-values}, is central to learning the Q-values from Equation \eqref{eq: Q-value definition} associated with state transitions $ (s, a, r, s') $ across a range of test $ \lambda $ values. These test values are selected as a random subset from `lambda\_grid', which encompasses a discretized set of $ \lambda $ values in the range of a problem-dependent $[-\lambda_{\max}, \lambda_{\max}]$, 
using 1000-point discretization.

During each iteration of the training process, the algorithm samples a batch of experiences from the memory. Each experience comprises the current state $ s $, the action taken $ a $, the reward received $ r $, the subsequent state $ s' $, and a boolean flag indicating the terminal status of $ s' $, i.e. whether $s'$ is the last state in an epoch, for a given individual project. For each experience, the algorithm computes the target Q-values for the state-action pair $ (s, a) $ using a random subset of $ \lambda $ values from `lambda\_grid'. This step involves evaluating the Q-value function for different levels of resource utilization and cost. By using a random subset of $\lambda$ values, the algorithm optimizes computation, reducing the number of evaluations needed for each update. Additionally, this approach helps to avoid overfitting by selecting different $\lambda$ points each time, ensuring that the model does not become too specialized to specific values of $\lambda$. The computation of the target Q-values $Q_{target}(s,a, \lambda)$ utilizes a target network, which is a lagged version of the primary neural network, to provide stable targets for learning \cite{van2016deep}.

Through this training process, the LPCA algorithm efficiently learns the Q-values for various state transitions under different levels of resource constraints, as dictated by the varying $ \lambda $ values. 

Having trained the neural network to generate accurate approximations of Equation \eqref{eq: Q-value definition}, we proceed with Algorithm \ref{alg: policy update} to compute the value function $ J(\mathbf{s}, \lambda) $ for a given state $\mathbf{s}$ as in Equation \eqref{eq: J-vector}. This computation involves evaluating $\sum_{i=1}^N \max_{a_i} Q(s_i, a_i, \lambda) $ for every $ \lambda $ within the discretized set `lambda\_grid'. 

Once this term is calculated, obtaining the optimal $\lambda^*$ is a one-dimensional convex optimization problem, as shown in Equation \eqref{eq: lambda optimal}.

\begin{algorithm}[t]
    \caption{Computation of Lagrange term $\lambda^*$}\label{alg: policy update}
    \begin{algorithmic}[1]
        \REQUIRE method
        \ENSURE Updated policy dictionary $\pi(\mathbf{s})$
    
        \STATE \textbf{function} PolicyDictUpdate(method)
        \FORALL{$\mathbf{s} \in \mathbf{S}$}
            \STATE q\_table $\leftarrow$ Zero Matrix of size [n\_lambda, N]
            \FOR{$i \in 1:N$}
                \STATE q\_table$[:,i] \leftarrow \max_{a_i} Q(s_i, a_i, \lambda), \forall \lambda \in \text{lambda\_grid}$
            \ENDFOR
            \STATE $J(\mathbf{s},\lambda) \leftarrow$ Compute value functions as (\ref{eq: J-vector})
            \STATE $\lambda^*(\mathbf{s})\leftarrow \arg\min_{\lambda} J(\mathbf{s},\lambda)$
            \IF{method = Evolution}
                \STATE $\mathbf{a}^* \leftarrow$ DifferentialEvolution($\mathbf{s}, \lambda^*(\mathbf{s}), a_{\max}$)
            \ELSE
                \STATE $\mathbf{a}^* \leftarrow$ Greedy($\mathbf{s}, \lambda^*(\mathbf{s}), a_{\max}, \delta$)
            \ENDIF
            \STATE $\pi(\mathbf{s}) \leftarrow \mathbf{a}^*$
        \ENDFOR
        \STATE \textbf{end function}
    \end{algorithmic}
\end{algorithm}


A key technical contribution of our work is how we explore the action space to solve the knapsack problem described in equation \eqref{eq: original constraint - forced}. This problem is challenging in neural networks due to the existence of many local minima, where traditional gradient optimization methods get stuck.

We propose two different strategies to explore this action space in order to make the best use of the available resources and select the best action based on our Q-value estimates. The first strategy, presented in Section \ref{sec:differential evolution}, is an evolutionary algorithm (LPCA-DE). It uses mechanisms similar to natural selection to iteratively search for the optimal solution, effectively avoiding local minima by exploring a wider range of solutions.

The second strategy, presented in Section \ref{sec: greedy}, is a greedy algorithm (LPCA-Greedy). It focuses on choosing the action based on the gradient of the Q-values with respect to the actions for each project, selecting the action that promises the highest increase in the Q-value per unit of resource expended.  This method is simpler and faster, and helps to quickly identify actions that increase payoff, even if it does not explore as widely.

\subsection{Differential Evolution Optimization (LPCA-DE)}\label{sec:differential evolution}
The first method (Algorithm \ref{alg:differential_evolution}) employs a differential evolution algorithm, renowned  for its effectiveness in identifying global optima and circumventing  local optima traps. This method is particularly adept at exploring the  search space comprehensively \cite{Das2011}.

A critical aspect of this approach is the integration of a penalty mechanism to ensure that action selection remains within resource constraints. Actions leading to resource utilization beyond the available limit are subjected to a significant penalty. This mechanism is in line with the role of the $\lambda$ term in the Q-value definition (see Equation \eqref{eq: Q-value definition}). 
Given the $\lambda c(a)$ term in Equation \eqref{eq: Q-value definition}, the derived optimal policy tends towards cost-effectiveness. However, it may not always coincide with the optimal policy of the original constrained problem (see Equation \eqref{eq: bellman coupled}) particularly if a higher action's benefit does not justify its cost in the relaxed problem, leading to potential underutilization of resources.
This leads to a policy that may not fully utilize the available resources as defined in Equation \eqref{eq: original constraint - forced}. 
To address this, we introduce an additional penalty, proportional to the amount of unused resources, into the differential evolution optimization problem. This modification guides the optimizer towards  actions that maximize resource usage, ensuring the algorithm not only pursues cost-effective solutions but also fully utilizes the available resources.
\begin{algorithm}[h]
    \caption{Action Selection through Differential Evolution Optimization}\label{alg:differential_evolution}
    \begin{algorithmic}[1]
        \REQUIRE State vector $\mathbf{s}$, fixed Lagrange multiplier $\lambda_{\text{fix}}$, maximum action $a_{\max}$
        \ENSURE Optimal actions maximizing Q-values under resource constraints
    
        \STATE \textbf{function} DifferentialEvolution($\mathbf{s}, \lambda_{\text{fix}}, a_{\max}$)
        \STATE $\text{Bounds} \leftarrow [0, a_{\max}]$
        \STATE \textbf{function} ObjectiveFunction($\mathbf{a}, \mathbf{s}, \lambda^*$)
        \STATE $Q_{\text{total}} \leftarrow \sum_{i=1}^{N} Q(s_i, a_i, \lambda^*)$
        \STATE $C_{\text{total}} \leftarrow \sum_{i=1}^{N} C(s_i, a_i)$
        \IF{$C_{\text{total}} > B$}
            \STATE $\text{Penalty} \leftarrow$ Large constant value
            \STATE $Q_{\text{total}} \leftarrow Q_{\text{total}} - \text{Penalty}$
        \ELSIF{$C_{\text{total}} < B$}
            \STATE $\text{Penalty} \leftarrow B - C_{\text{total}}$
            \STATE $Q_{\text{total}} \leftarrow Q_{\text{total}} - \text{Penalty}$
        \ENDIF
        \RETURN $-Q_{\text{total}}$
        \STATE \textbf{end function}
        \STATE $\mathbf{a}^* \leftarrow \text{Apply Differential Evolution optimization with }(ObjectiveFunction, Bounds)$
        \RETURN $\mathbf{a}^*$
        \STATE \textbf{end function}
    \end{algorithmic}
\end{algorithm}

\subsection{Greedy Optimization Strategy (LPCA-Greedy)}\label{sec: greedy}
The second method (Algorithm \ref{alg:greedy_selection}) is a greedy optimization strategy. This approach is characterized by its iterative process of evaluating the gradient of the Q-values with respect to the actions for each project and then allocating resources to the project with the highest gradient. The process continues until all resources are exhausted.

This strategy prioritizes complete resource utilization, assigning resources to the projects that promise the highest increase in the Q-value per unit of resource expended. Unlike the differential evolution method, which searches for an optimal policy and then adjusts for resource utilization, the greedy approach begins with the premise of full resource allocation and does so in a manner that maximizes the benefit derived from each project.

The choice between these methods can be guided by the specific characteristics of the problem at hand, such as the nature of the resource constraints and the desired balance between resource utilization and reward maximization.

\begin{algorithm}[h]
    \caption{Greedy Action Selection for Continuous MDP}\label{alg:greedy_selection}
    \begin{algorithmic}[1]
        \REQUIRE State $\mathbf{s}$, $\lambda_{\text{fix}}$, max action $a_{\text{max}}$, increment $\delta$
        \ENSURE Optimal actions maximizing Q-values, maximum action $a_{\max}$
        
        \STATE \textbf{function} Greedy($\mathbf{s}, \lambda_{\text{fix}}, a_{\text{max}}, \delta$)
        \STATE Initialize action vector $\mathbf{a}$ to zeros, $B_{\text{remaining}} = B$
        \WHILE{$B_{\text{remaining}} > 0$}
            \STATE $i \leftarrow \arg\max_i \frac{\partial Q}{\partial a_i}$
            \STATE $a_i \leftarrow a_i + \delta$, ensure $a_i \leq a_{\max}$
            \STATE $B_{\text{remaining}} \leftarrow B - \sum_{i=1}^{N} c(s_i, a_i)$
        \ENDWHILE
        \RETURN $\mathbf{a}$
        \STATE \textbf{end function}
    \end{algorithmic}
\end{algorithm}


\section{Experimental Results}\label{sec: experiments}
To evaluate the effectiveness of our algorithms, we rely on measuring the average discounted rewards that their policies yield. Given a discount factor of $\gamma=0.9$, we examine the rewards that each algorithm's policy yields over 
$t \in [0,50]$ iterations,
starting from every possible state in our problem space. The evaluation process involves computing the discounted sum of the rewards using the equation 
$$R = \sum_{t=0}^{50} \sum_{i=1}^N \gamma^t r(s_i(t), a_i(t)),$$ 
where $r(s_i(t), a_i(t))$ represents the reward received at time $t$ for being in state $s_i(t)$ and taking action $a_i(t)$, for each MDP $i$. To ensure statistical robustness and to derive confidence intervals for our performance metrics, we repeat this evaluation 100 times. The results are shown in our figures, with the mean performance represented by bold lines and the confidence intervals represented by the shaded areas surrounding these lines.

Our experimental framework encompasses three distinct types of problems: Type A and Type B, each representing a continuous action version of challenges similar to those discussed in \cite{Killian2021}, and the speed scaling problem inspired from \cite{Wierman2009}. Types A and B feature two states per project with $a \in [0,2]$, with a reward function $ R(s)=s $ and a cost function $ C(a)=a $. The key difference between Type A and Type B lies in their transition probability matrices:

$$
\begin{aligned}
&& P_A(a) =& \begin{pmatrix}

0.02a^2 - 0.09 a + 0.8 & -0.02a^2+0.09a+0.2
\\
0.75 e^{-0.947a} & 1-0.75e^{-0.947a}

\end{pmatrix}
\\
&&P_B(a) = &\begin{pmatrix}

0.95e^{-2.235a} & 1-0.95e^{-2.235a}
\\
0.3347e^{-1.609a} & 1-0.3347e^{-1.609a}
\end{pmatrix}.
\end{aligned}
$$

Additionally, we introduce a mixed environment where half of the projects follow the transition probabilities of Type A and the other half those of Type B. 

The speed scaling environment involves projects with six states, and $a \in [0,2]$. 
We apply the uniformization technique \cite{Puterman2014} to construct an equivalent discrete time version of the continuous time problem.
The transition probabilities are given by

$$
{
    P(a) = 
    \begin{pmatrix}
    1 - \frac{\alpha}{\nu} & \frac{\alpha}{\nu} & 0 & 0 & 0 & 0 \\
    \frac{\mu_a}{\nu} & 1 - \frac{\alpha + \mu_a}{\nu} & \frac{\alpha}{\nu} & 0 & 0 & 0 \\
    0 & \frac{\mu_a}{\nu} & 1 - \frac{\alpha + \mu_a}{\nu} & \frac{\alpha}{\nu} & 0 & 0 \\
    0 & 0 & \frac{\mu_a}{\nu} & 1 - \frac{\alpha + \mu_a}{\nu} & \frac{\alpha}{\nu} & 0 \\
    0 & 0 & 0 & \frac{\mu_a}{\nu} & 1 - \frac{\alpha + \mu_a}{\nu} & \frac{\alpha}{\nu} \\
    0 & 0 & 0 & 0 & \frac{\mu_a}{\nu} & 1 - \frac{\mu_a}{\nu}
    \end{pmatrix},
}
$$
where $\alpha=0.9$ is the arrival rate, $ \mu(a) = \sqrt{a} $ is the controlled departure rate, $ \nu = \max_a (\alpha + \mu(a))$ is the normalization factor, $ \beta $ is the continuous discount factor related to the discrete factor $ \gamma $ as $\beta= \frac{\nu}{\gamma} - \nu $. The reward function is defined as:
$$
R(s) = \frac{-s}{\nu+\beta}+ \frac{C_r}{\nu+\beta} = 
\begin{cases}
\frac{-s}{\nu+\beta} & \text{if } s< s_{\max} \\
\frac{-s_{\max}-10}{\nu+\beta} & \text{if }s = s_{\max}
\end{cases}
$$
where $C_r = -10$ is the rejection cost that occurs in the final state $s_{\max}$. The cost function is defined as $ C(s,a) = \frac{a}{\nu + \beta} $ if $ s > 0 $, otherwise 0.

For Types A and B, we conducted experiments with both 4 projects with 2 units of resources and 6 projects with 4 units of  resources. The mixed environment, combining Types A and B, was tested  with 6 projects and 4 units of resources. The Speed Scaling experiment involved 4 projects with 1.5 units of  resources, equivalent to fully activating two of the four projects.

To benchmark our algorithm, we choose DDPG (Deep Deterministic Policy Gradient) \cite{Lillicrap2019} augmented with OptLayer \cite{Pham2018} as the baseline. OptLayer enhances DDPG by incorporating a constraint optimization layer in the actor network, enabling the generation of actions that respect the constraints outlined in the original problem formulation (Equations \eqref{eq: bellman coupled} and \eqref{eq: original constraint - forced}).

In addition to this, we have benchmarked Whittle's index heuristic for continuous actions. These indices are computed through the algorithm proposed by \cite{weber2007comments} for discrete multi-action ($a \in [0, 1, 2, \dots]$) and adapted for an arbitrary discretization $\delta_a$ of the action ($a \in [0, \delta_a, 2\delta_a, \dots]$). For an approximation of a fully continuous action, we use a discretization of $\delta_a = 0.001$, leading to a total of 2001 possible actions. Due to the large amount of indices to compute, a tabular learning algorithm for those indices would not be feasible.

\begin{figure}[t]
    \centering
    \includegraphics[width=0.49\linewidth]{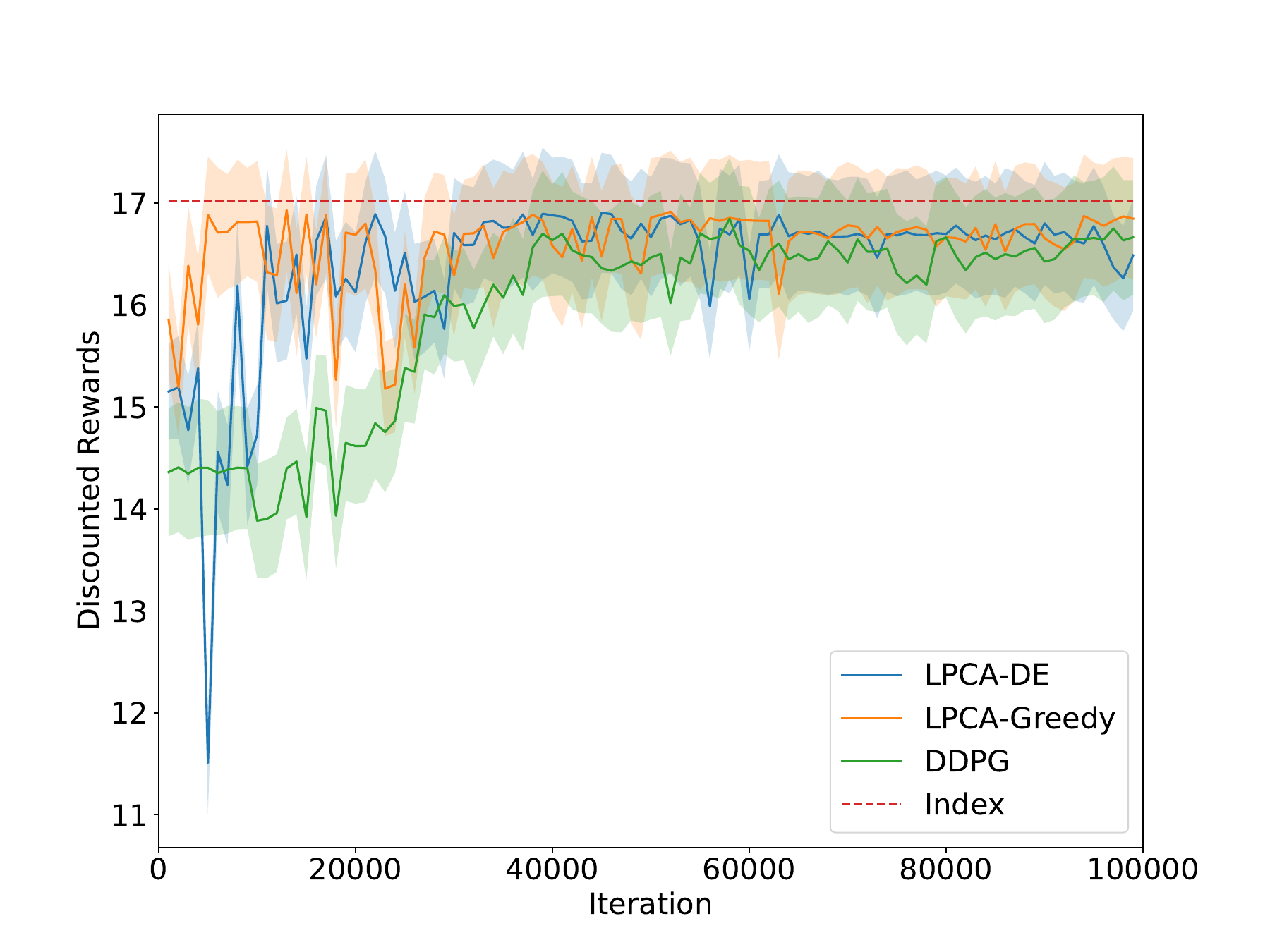}
    \hspace{0.001\linewidth}
    \includegraphics[width=0.49\linewidth]{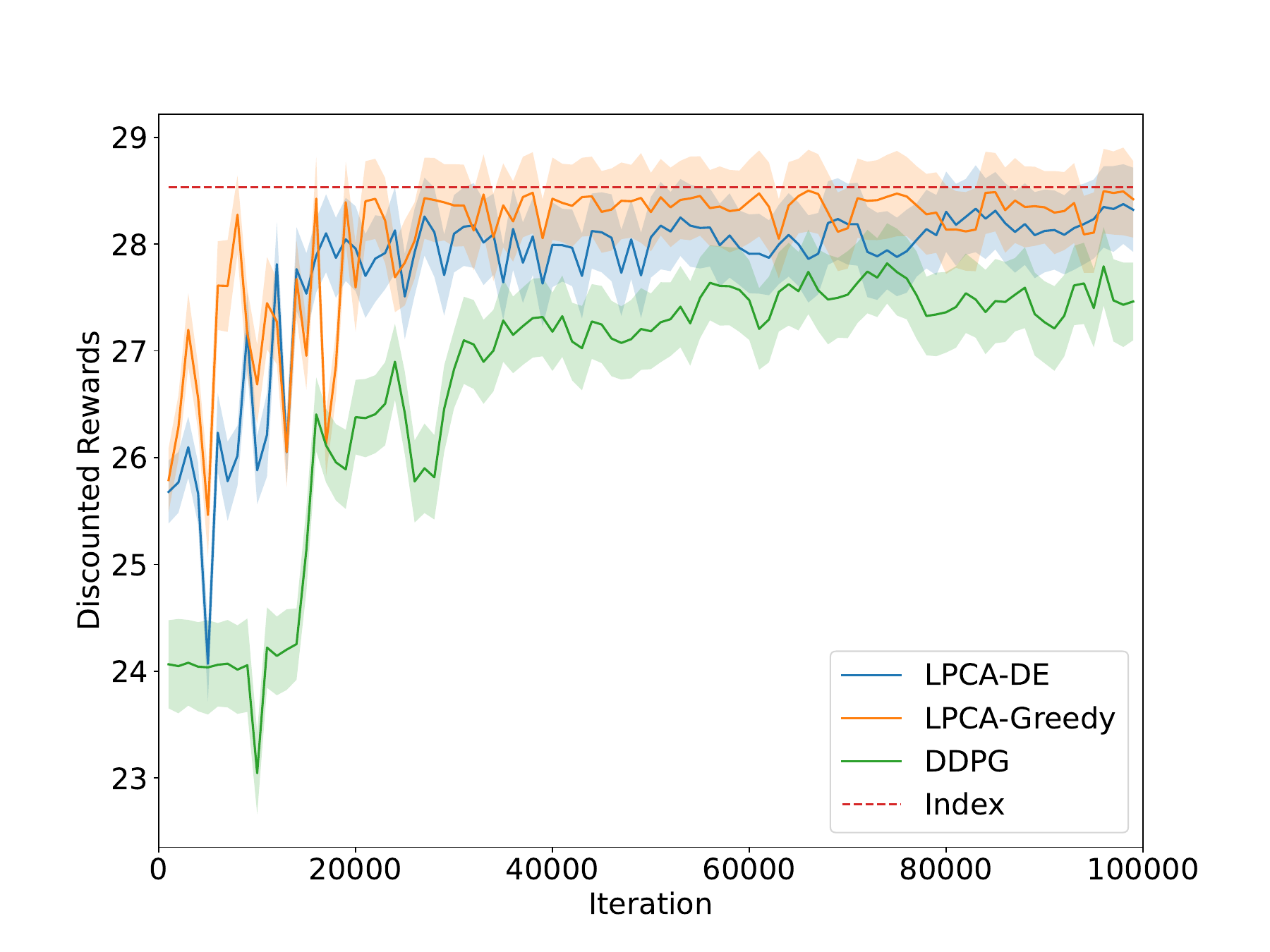}
    \caption{Experimental results for Type A environment: (Left) 4 projects and 2 units of resources, (Right) 6 projects and 4 units of resources.}
    \label{fig:typeA}
\end{figure}

\begin{figure}[h]
    \centering
    \includegraphics[width=0.49\linewidth]{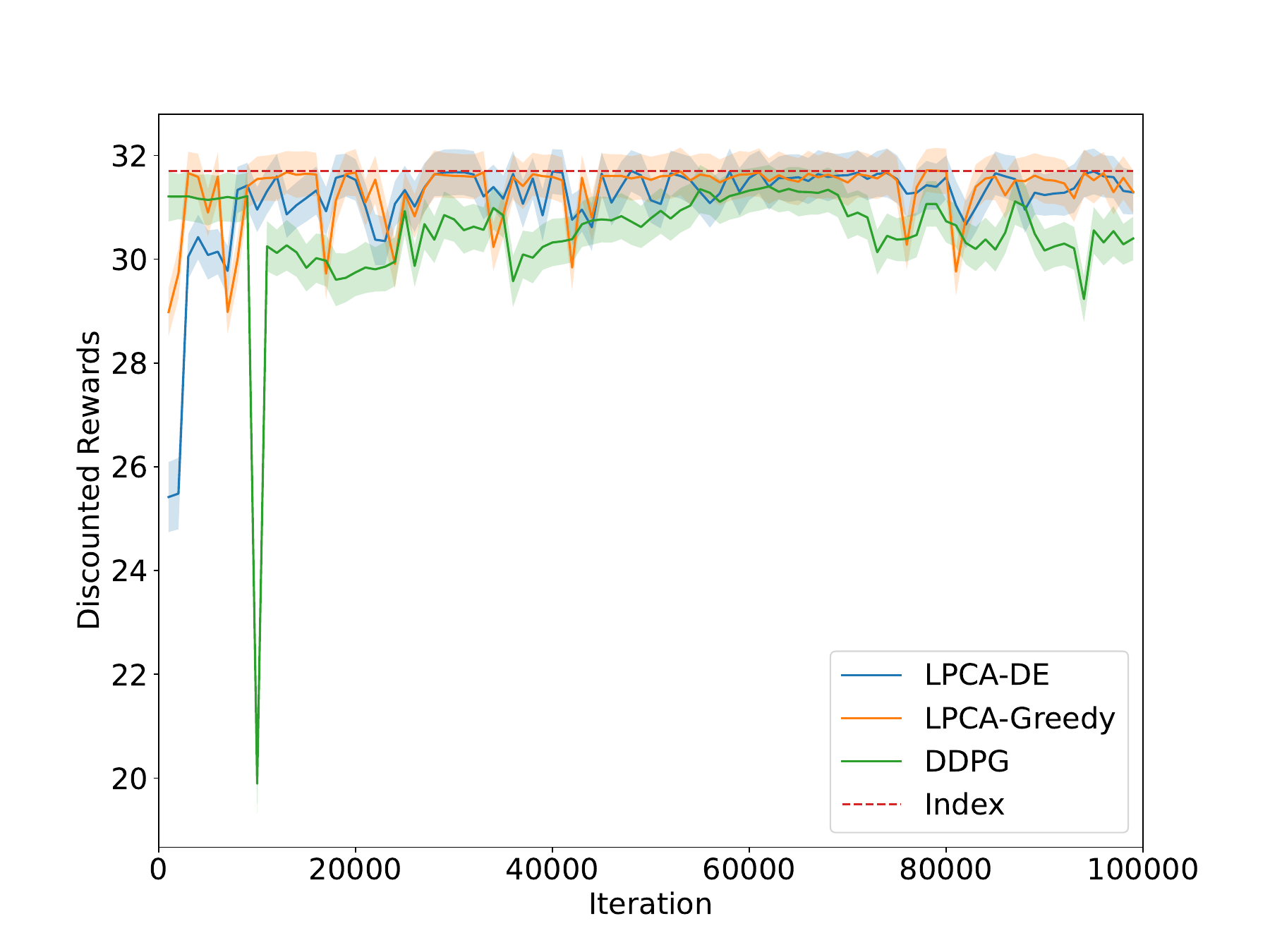}
    \hspace{0.001\linewidth}
    \includegraphics[width=0.49\linewidth]{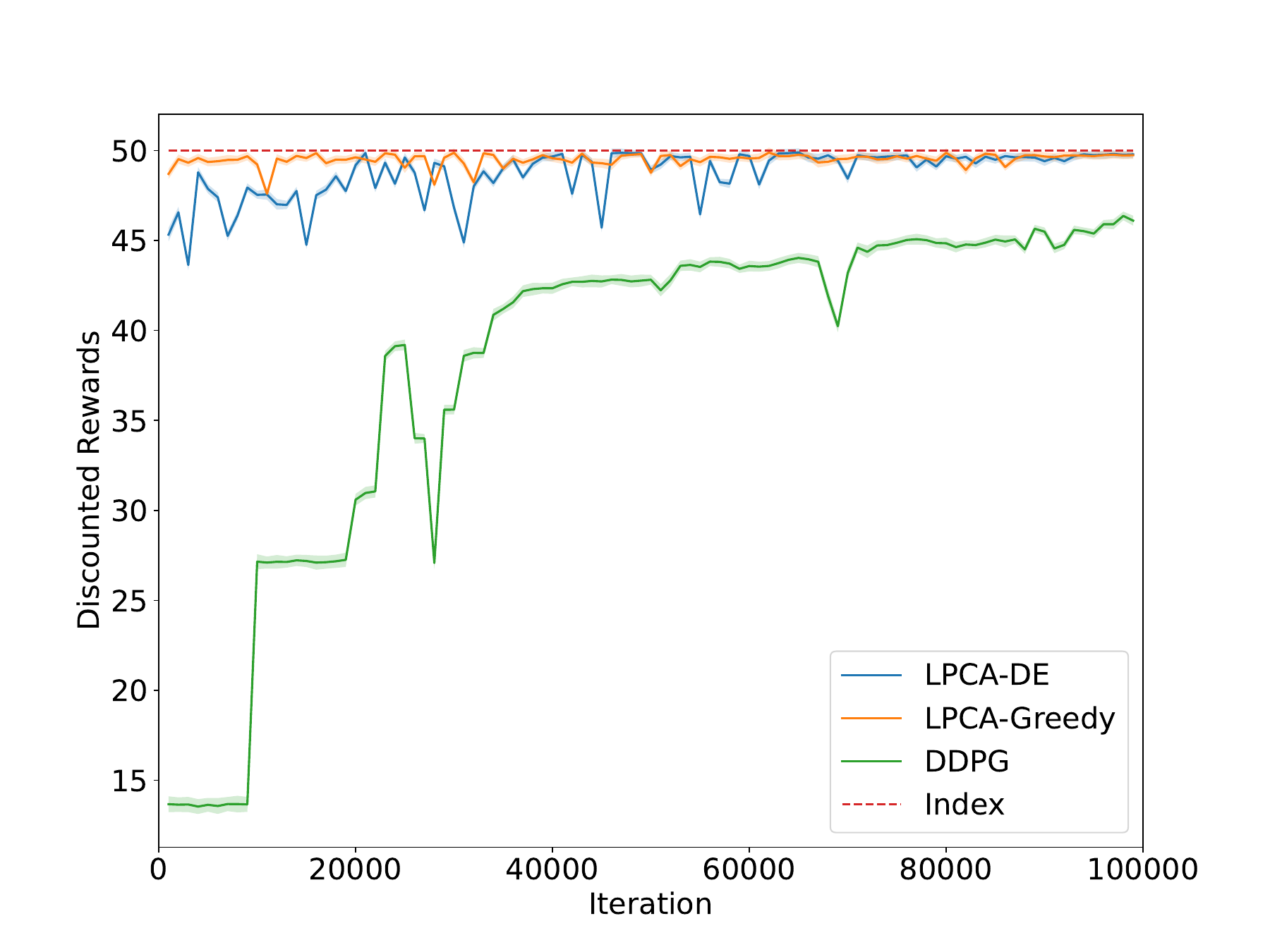}
    \caption{Experimental results for Type B environment: (Left) 4 projects and 2 units of resources, (Right) 6 projects and 4 units of resources.}
    \label{fig:typeB}
\end{figure}

\begin{figure}[h]
    \centering
    \includegraphics[width=0.49\linewidth]{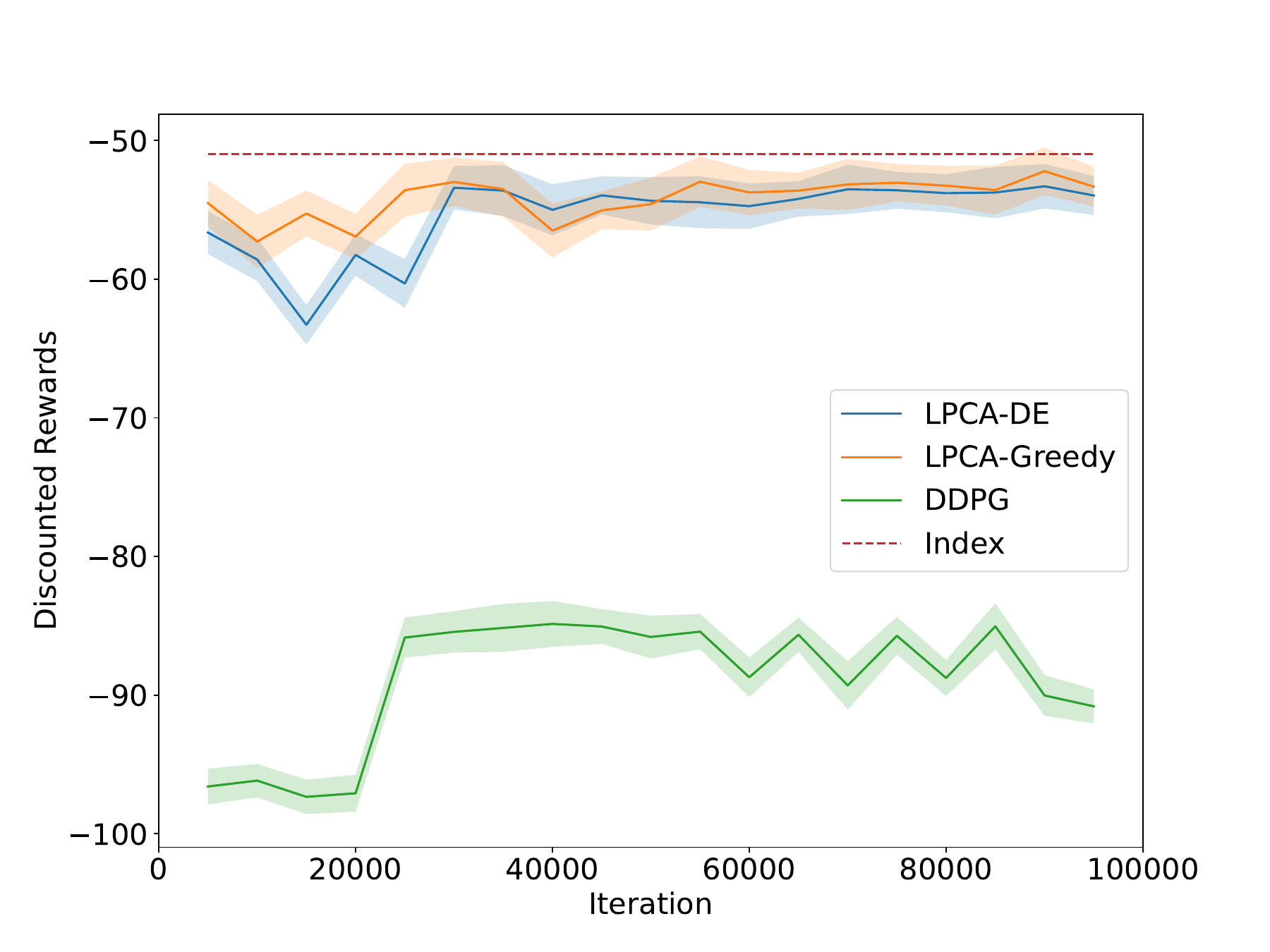}
    \hspace{0.001\linewidth}
    \includegraphics[width=0.49\linewidth]{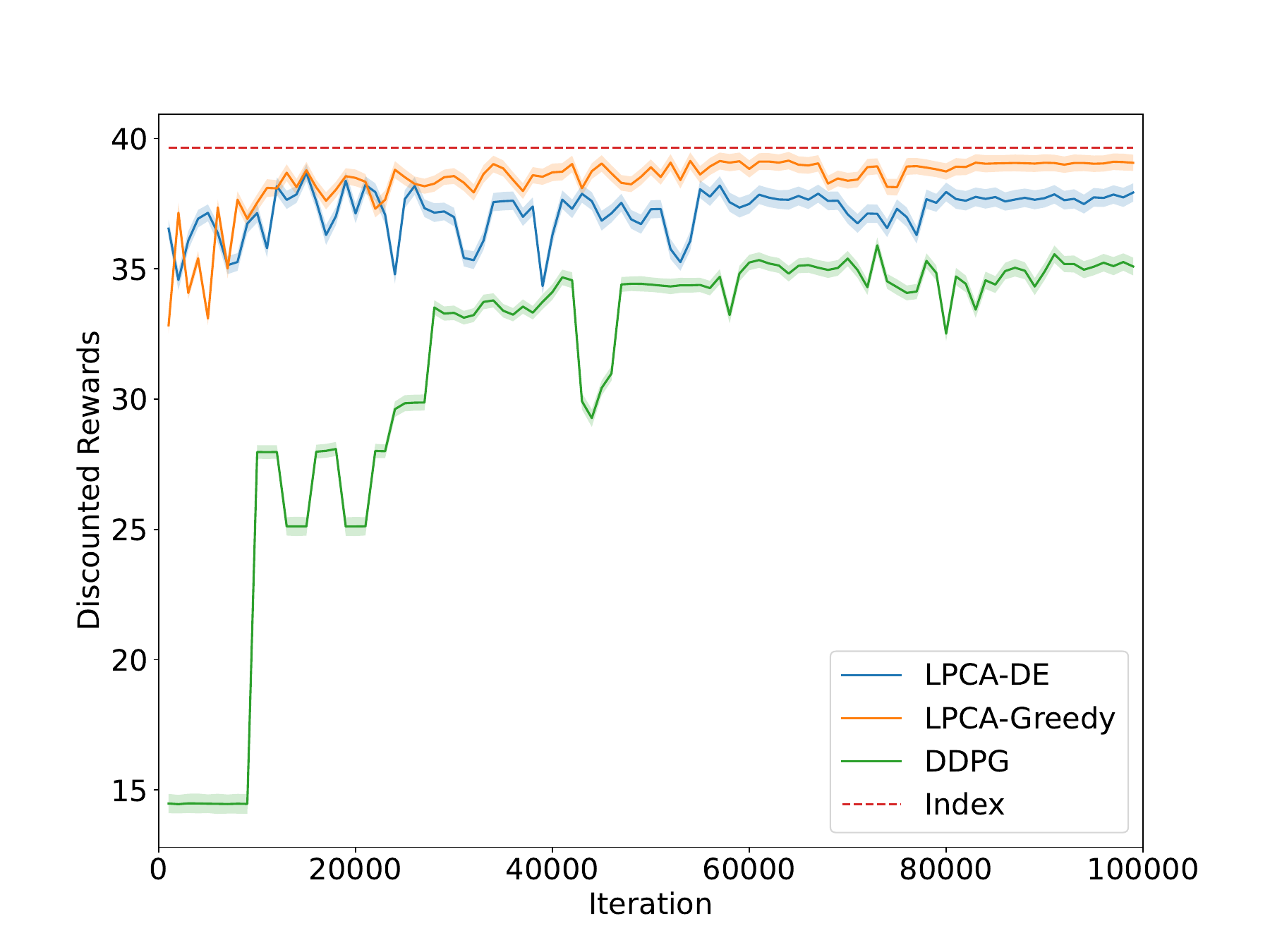}
    \caption{(Left) Speed Scaling with  4 projects and 1.5 units of resources, (Right) Mixed Type A and B environments with 6 projects and 4 units of resources.}
    \label{fig:mixed}
\end{figure}

In the 4 projects and 2 resources configuration, both LPCA-DE and LPCA-Greedy demonstrated a clear advantage over DDPG, particularly in Type B environment (Figure \ref{fig:typeB} left), where the gap between the performance of both versions of LPCA and DDPG is larger and both LPCA algorithms converge to the Whittle Index policy performance. In Figure \ref{fig:typeA} left, although DDPG achieves a similar level of performance to LPCA, the latter converges to a performance level similar to Whittle indices' much faster, while DDPG takes around 40000 iterations. 

This gap widened significantly in the 6 projects and 4 resources setting. In Type A (Figure \ref{fig:typeA} right), the optimality gap between both versions of LPCA and DDPG widens. A similar pattern shows in Type B (Figure \ref{fig:typeB} right), with DDPG having subpar performance. 
In the mixed environment (Figure \ref{fig:mixed} right), DDPG's performance reflects the issues observed in the previous scenarios. On the other hand, LPCA-DE and specially LPCA-Greedy are able to obtain a better policy, close to the Whittle index policy performance. 

In the Speed Scaling experiment (Figure \ref{fig:mixed} left), the performance of both LPCA-DE and LPCA-Greedy algorithms converges to a similar performance to the Whittle index policy, while DDPG's performance lags behind.


Overall, the LPCA algorithms consistently outperformed DDPG with OptLayer across various settings and environments. Notably, LPCA's superiority became increasingly pronounced in more complex scenarios involving a greater number of processes and limited resources.


\section{Conclusion}\label{sec: conclusion}
In this study, we introduced the LPCA (Lagrange Policy for Continuous Actions) algorithm, a reinforcement learning approach for weakly coupled MDPs with continuous actions and resource constraints. Our experimental results demonstrate that LPCA, in both its Differential Evolution (DE) and Greedy variants, consistently outperforms the DDPG algorithm augmented with OptLayer across various scenarios. 
Notably, LPCA exhibits superior scalability with an increasing number of projects.

As a direction for future research, we aim to test the LPCA algorithm in larger-scale environments featuring more states per projects. This expansion will allow us to further evaluate LPCA's scalability and effectiveness in even more complex and dynamic settings, potentially broadening its applicability to a wider array of practical problems in operations research and beyond. The exploration of LPCA's performance in these extended scenarios is expected to yield valuable insights into its capabilities and limitations, guiding future enhancements and adaptations of the algorithm.

\begin{credits}
\subsubsection{\ackname} F. Robledo has received funding from the Department of Education of the Basque Government through the Consolidated Research Group MATHMODE (IT1456-22). Research partially supported by the French ``Agence Nationale de la Recherche (ANR)'' through the project ANR-22-CE25-0013-02 (ANR EPLER) and DST-Inria Cefipra project LION.

This paper has been accepted in ASMTA 2024 Conference


\subsubsection{\discintname}
The authors have no competing interests to declare that are
relevant to the content of this article.
\end{credits}
%
%
%
%

\bibliography{bib_new}
\bibliographystyle{splncs04}





\end{document}